\documentclass[english,american,prr,twocolumn,amsmath,amssymb,superscriptaddress,nofootinbib,floatfix,longbibliography]{revtex4-2}

\usepackage{graphicx}
\usepackage{float}
\usepackage{stfloats}
\usepackage{natbib}
\usepackage{amssymb,amsfonts,amsmath}
\usepackage[english]{babel}             
\usepackage[T1]{fontenc}                
\usepackage[utf8]{inputenc}             
\usepackage{mathtools}                  
\usepackage{booktabs}                   
\usepackage{units}
\usepackage[colorlinks=true, linkcolor=black, citecolor=black, urlcolor=blue]{hyperref}
\usepackage[title]{appendix}
\usepackage{prettyref}
\usepackage[colorinlistoftodos, textsize=small]{todonotes}
\usepackage{afterpage}
\usepackage[morefloats=24]{morefloats}
\usepackage{wrapfig}

\newrefformat{fig}{\hyperref[#1]{Fig.~\ref*{#1}}}
\newrefformat{tab}{\hyperref[#1]{Table~\ref*{#1}}}
\newrefformat{eq}{\hyperref[#1]{Eq.~(\ref*{#1})}}
\newrefformat{sec}{\hyperref[#1]{Section~\ref*{#1}}}
\newrefformat{subsec}{\hyperref[#1]{Section~\ref*{#1}}}
\newrefformat{app}{\hyperref[#1]{Appendix~\ref*{#1}}}
\newrefformat{cha}{\hyperref[#1]{Chapter~\ref{#1}}}

\usepackage[colorinlistoftodos, textsize=small]{todonotes}

\newif\ifcomments
\commentsfalse  

\ifcomments
  \newcommand{\todocomment}[1]{\todo{#1}}
\else
  \newcommand{\todocomment}[1]{}  
\fi

\usepackage{xcolor}

\newcommand{\R}{\mathbb{R}}

\usepackage{fancyhdr}
\begin{document}

\title{Symmetry and Generalisation in Neural Approximations of Renormalisation Transformations}

\author{Cassidy Ashworth}
\email{cga28@cam.ac.uk}
\author{Pietro Liò}
\email{p1219@cam.ac.uk}
\affiliation{University of Cambridge, Cambridge, UK}
\author{Francesco Caso}
\email{francesco.caso@uniroma1.it}
\affiliation{University of Cambridge, Cambridge, UK}
\affiliation{University of Rome, La Sapienza, Rome, Italy}

\begin{abstract}
Deep learning models have proven enormously successful at using multiple layers of representation to learn relevant features of structured data. Encoding physical symmetries into these models can improve performance on difficult tasks, and recent work has motivated the principle of parameter symmetry breaking and restoration as a unifying mechanism underlying their hierarchical learning dynamics. We evaluate the role of parameter symmetry and network expressivity in the generalisation behaviour of neural networks when learning a real-space renormalisation group (RG) transformation, using the central limit theorem (CLT) as a test case map. We consider simple multilayer perceptrons (MLPs) and graph neural networks (GNNs), and vary weight symmetries and activation functions across architectures. Our results reveal a competition between symmetry constraints and expressivity, with overly complex or overconstrained models generalising poorly. We analytically demonstrate this poor generalisation behaviour for certain constrained MLP architectures by recasting the CLT as a cumulant recursion relation and making use of an established framework to propagate cumulants through MLPs. We also empirically validate an extension of this framework from MLPs to GNNs, elucidating the internal information processing performed by these more complex models. These findings offer new insight into the learning dynamics of symmetric networks and their limitations in modelling structured physical transformations.

\end{abstract}
\maketitle


\section{Introduction\label{sec:introduction}}

The renormalisation group (RG) is a cornerstone of modern theoretical physics, used in areas ranging from critical phenomena in phase transitions to the Standard Model \cite{peskin2018introduction, wilson_renormalization_1971}. The RG method often proceeds à la Wilson, progressively eliminating high frequency modes to obtain large-scale system properties \cite{wilson_renormalization_1983, wilson_renormalization_1971, kadanoff_scaling_1966}. However, from a probabilistic viewpoint the RG can be interpreted as transforming distributions over physical degrees of freedom \cite{jona-lasinio_renormalization_2001}. Order parameters are interpreted as random fields and the relevant limit distributions are reproduced upon coarse-graining, providing a clear interpretation of statistical universality. This focus on distributional transformations establishes a clear conceptual link between RG and neural networks, which can similarly be characterised as a mapping between input and output data distributions \cite{fischer_decomposing_2022}. This link has been the subject of a number of recent works in which formal connections have been drawn \cite{beny_deep_2013, mehta_exact_2014, iso_scale-invariant_2018, de_mello_koch_is_2020, lenggenhager_optimal_2020}.

Neural networks have proven incredibly successful at solving a wide range of tasks, from molecule prediction \cite{duvenaud_convolutional_2015, kearnes_molecular_2016, gilmer_neural_2017} and material science \cite{reiser_graph_2022, merchant_scaling_2023} to modelling physical interactions \cite{sanchez-gonzalez_learning_2020} and improving PDE solvers \cite{brandstetter_message_2023, mayr_boundary_2023, rotman_semi-supervised_2023, mizera_scattering_2023, benitez_neural_2025}. While there has been considerable progress in theoretical understanding \cite{bahri_statistical_2020, lin_why_2017}, their precise decision-making mechanisms remain mostly opaque \cite{fischer_decomposing_2022, erdmenger_towards_2022}. A better understanding of the functional principles that underlie their success would allow networks to be more easily tailored to specific tasks. A promising avenue for improved interpretability and generalisation is the principled encoding of physical symmetries into network architectures \cite{wang_incorporating_2021, lin_why_2017}, with the added reciprocal benefit that better deep learning models can offer useful physical insight \cite{marchand_wavelet_2023, hashimoto_adscft_2019}.

While prior work has focused on applying RG theory to directly improve model performance \cite{caso_renormalized_2025}, here we shift the focus to instead improving an understanding of \textit{how} neural networks learn. We evaluate the role that parameter symmetry and network expressivity play in the generalisation of neural models learning physical transformations, using the RG-inspired central limit theorem (CLT) as a test case map. We empirically find a tension between symmetry constraints and network expressivity, and obtain analytical results to this end for certain simple Multilayer Perceptrons (MLPs) by recasting the CLT as a transformation of cumulants and theoretically propagating these cumulants through MLPs. We extend this propagation framework to graph neural networks (GNNs) and empirically validate our findings, improving the interpretability of these models.

\section{Theory\label{sec:Theoretical-Background}}

\subsection{CLT as an RG Transformation}
\label{subsec:RG_CLT}

Following Jona-Lasinio \cite{jona-lasinio_renormalization_2001}, the CLT asserts the following. Let $\xi_1, \xi_2, \ldots, \xi_n, \ldots$ be a sequence of independent identically distributed (i.i.d.) random variables with finite variance $\sigma^2 = \mathbb{E}(\xi_i - \mathbb{E}(\xi_i))^2$, with $\mathbb{E}$ indicating an expectation with respect to their common distribution. Then
\begin{equation}
\frac{\sum_{i}(\xi_i - \mathbb{E}(\xi_i))}{\sigma \sqrt{n}} \xrightarrow{n \to \infty} N(0,1),
\end{equation}
with $N(0,1)$ the normal centered distribution of variance 1.

The system can be viewed as a real space decimation transformation on a one-dimensional Ising model. Consider $\xi_i$ as spins on a one-dimensional lattice $\mathbb{Z}$ and define $\zeta_n^1 = 2^{-n/2} \sum_{i=1}^{2^n} \xi_i$ and $\zeta_n^2 = 2^{-n/2} \sum_{i=2^n + 1}^{2^{n+1}} \xi_i$. Then
\begin{equation}
\zeta_{n+1} = \frac{1}{\sqrt{2}} \left( \zeta_n^1 + \zeta_n^2 \right). \label{eq:clt decimation}
\end{equation}
\prettyref{eq:clt decimation} is the RG transformation that networks learn in subsequent generalisation experiments.

The recursion relation for the corresponding distributions is
\begin{equation}
p_{n+1}(x) = \sqrt{2} \int dy\, p_n(\sqrt{2}x - y)\, p_n(y) = (\mathcal{R} p_n)(x).\label{eq:distrib_recursion}
\end{equation}
By considering the cumulant generating function, we recast \prettyref{eq:distrib_recursion} in terms of $r$th order cumulants at RG step $n$ as
\begin{equation}
    \kappa_r^{(n+1)} = 2^{1-r/2} \kappa_r^{(n)} \quad \text{for } r \geq 1.\label{eq:laplace}
\end{equation}
\prettyref{eq:laplace} makes clear that all cumulants with $r > 2$ are suppressed under the CLT. This result is used in \prettyref{subsec:analytical weights} to obtain analytical solutions for symmetrically-constrained network weights.

By substitution it can be checked that the family of Gaussians
\begin{equation}
p_{G, \sigma}(x) = \frac{1}{\sqrt{2\pi \sigma^2}} e^{-x^2 / 2\sigma^2} \label{gaussian_fixed}
\end{equation}
\noindent
are fixed points of \prettyref{eq:distrib_recursion}, preserving normalisation, mean, and variance, for centred distributions. More detail is contained in \prettyref{app:CLT recursion}.

\subsection{Multilayer Perceptrons (MLPs)}
\subsubsection{Network Architecture}
\label{subsubsec:MLP architecture}

Here we outline the structure of MLPs, fully connected neural networks with $N_{L}$ neurons in each layer $L$. Layers consist of an affine transformation 
\begin{equation}
z_{i}^{l}=\sum_{j=1}^{N_{l-1}}\,W_{ij}^{l}\,y_{j}^{l-1}+b_{i}^{l}\label{eq:pre_activations}
\end{equation}
parameterized by a weight matrix $W^{l}\in\R^{N_{l}\times N_{l-1}}$
and bias vector $b^{l}\in\R^{N_{l}}$. A nonlinear activation function $\phi$ is then applied elementwise such that the full layer is defined as
\begin{equation}
y_{i}^{l}=\phi\big(z_{i}^{l}\big)=\phi\left(\sum_{j=1}^{N_{l-1}}\,W_{ij}^{l}\,y_{j}^{l-1}+b_{i}^{l}\right).\label{eq:post_activations}
\end{equation}
Input data of dimension $N_0$ is denoted as $y^{0}=x\in\R^{N_{0}}$. Iterating over layers defines the network mapping $y=g(x;\theta)$ for parameters $\theta:=\{W^{l},b^{l}\}_{l=1,\dots,L+1}$, with the output $y_{i}=z_{i}^{L+1}, \; y\in\R^{d_{\text{out}}}$ given by a final linear readout layer.

\begin{figure}[h]
    \centering
    \includegraphics[width=\linewidth]{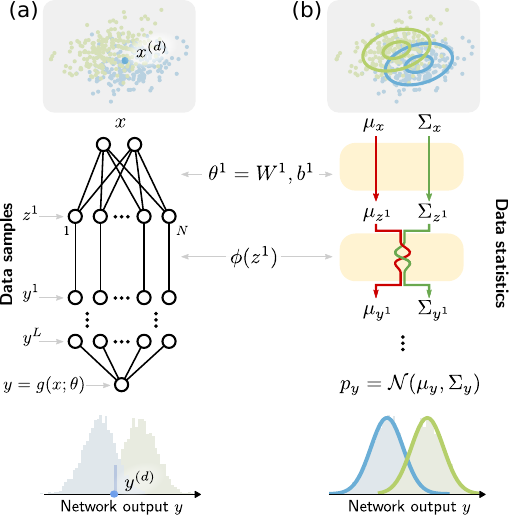}
    \caption{(a) Network analysis based on data samples considers a direct mapping between samples $x$ and outputs $y$ comprised of affine and nonlinear transformations. (b) The alternative statistical viewpoint considers how networks transform the entire data distribution $p(x)$. Affine layers propagate cumulants independently but the nonlinearity mixes cumulants of different orders $n$. A Gaussian input with only non-zero mean $\mu$ and covariance $\Sigma$ is shown. Figure taken from \cite{fischer_decomposing_2022}.}
    \label{fig:MLP_architecture}
\end{figure}

\subsubsection{Decomposing MLPs as Mappings of Correlation Functions}
\label{subsubsec:MLP cumulants}

Following Fischer et al \cite{fischer_decomposing_2022}, here we outline how cumulants propagate through MLP networks. For more detail, see \prettyref{app:statistical model}.

Cumulants are the preferred distributional parameterisation as they are additive under addition of independent variables, leading to simpler expressions for their propagation. Given the network mapping $g : x \mapsto y$ we could relate the cumulant generating function of the outputs $y$ to that of the inputs $x$ as
\begin{align}
\mathcal{W}_{y|\theta}(j) &= \ln \langle \exp(j^\top y) \rangle_{y|\theta} \\
&= \ln \langle \exp(j^\top g(x; \theta)) \rangle_{x}. \label{eq:generating function output}
\end{align}
The output cumulant of order $n$ is defined as the derivative
\begin{equation}
    G_{y|\theta}^{(n)} = \left. \frac{d^n \mathcal{W}_{y|\theta}(j)}{dj^n} \right|_{j=0}.
\end{equation}
In principle, evaluating \prettyref{eq:generating function output} would yield a direct relation between input and output cumulants. However, while the nonlinear activation $\phi$ is critical for network expressivity \cite{cybenko_approximation_1989, hornik_multilayer_1989}, it makes direct evaluation of $\mathcal{W}_{y|\theta}(j)$ difficult. Instead, we focus on tracking the layerwise cumulant propagation.

An affine linear transformation of the form $y_i = W_{ij}x_j + b_i$ relates the preactivations $z^{l}$ to the postactivations of the previous layer $y^{l-1}$. The corresponding cumulant generating function transformation is
\begin{align}
    \mathcal{W}_{z^{l}}(j) &= \ln \langle \exp(j^\top z^{l}) \rangle_{z^l} \nonumber \\
    &= \ln \langle \exp(j^\top W^{l} y^{l-1} + j^\top b^l) \rangle_{y^{l-1}} \nonumber \\
    &= \mathcal{W}_{y^{l-1}}((W^l)^\top j) + j^\top b^l, \label{eq:affine transf_re}
\end{align}
which gives
\begin{equation}
    G_{z^{l}}^{(1)} = W^l G_{y^{l-1}}^{(1)} + b^l \label{eq:affine_mean_re}
\end{equation}
for the first order cumulant ($n=1$), and
\begin{equation}
    G_{z^{l},i_1,\dots,i_n}^{(n)} = \sum_{s_1,\dots,s_n} W^l_{i_1 s_1} \cdots W^l_{i_n s_n} \, G_{y^{l-1},s_1,\dots,s_n}^{(n)} \label{eq:affine_higher_re}
\end{equation}
for $n \geq 2$.

\vspace{1em}
We can write the equivalent transformation across the nonlinear activation function $\phi$ within each layer $l$ as
\begin{align}
    \mathcal{W}_{y^{l}}(j) &= \ln \langle \exp(j^\top y^{l}) \rangle_{y^{l}} \nonumber \\
&= \ln \langle \exp(j^\top \phi(z^{l})) \rangle_{z^{l}}. \label{eq:activation_MLP}
\end{align}
In general, \prettyref{eq:activation_MLP} cannot be evaluated exactly \cite{fischer_decomposing_2022}. However, perturbative techniques exist to reconstruct postactivation cumulants as functions of preactivation equivalents, $G_{y^{l}}^{(n)} = f_{n}(\{ G_{z^{l}}^{(m)} \})$ \cite{helias_statistical_2020}. This mixing of cumulants represents the information transfer between layers.

\prettyref{eq:affine_mean_re}, \prettyref{eq:affine_higher_re}, and the relevant interaction functions $f_{n}(\{ G_{z^{l}}^{(m)} \})$ can be composed to define a \text{statistical model} corresponding any given network as
\begin{equation}
    g_{\text{stat}} : \left( \left\{ G_{x}^{(m)} \right\}_{m}, \theta, \phi \right) \mapsto p(y).
\end{equation}
$p(y)$ is characterised by the propagated cumulants
\begin{align}
G_y^{(n)} &= W^{L+1} \left( f_{n} \left( \cdots \{ G_{x}^{(m)} \}_{m}  \cdots\right) \right) + b^{L+1} \nonumber\\
&=: g_{n} \left( \{ G_{x}^{(m)} \}_{m}; \theta, \phi \right).
\end{align}
This framework characterises MLPs as mappings of data distributions.

\subsection{Graph Neural Networks (GNNs)}
\subsubsection{Network Architecture}
\label{subsubsec:GNN_architecture}
We can define a general graph $\mathcal{G} = (\mathcal{V, E})$ with nodes $v_{i} \in \mathcal{V}$, edges $e_{ij} \in \mathcal{E}$, $N$-dimensional node feature vectors $h_i \in \mathbb{R}^N$, and denote the set of nodes neighbouring $v_i$ as $\mathcal{N}(i)$. GNNs use a message passing framework whereby each $h_i$ is iteratively updated using information aggregated across its neighbourhood $\mathcal{N}(u)$ \cite{hamilton2020graph}. The different iterations are known as the layers of a GNN. The basic update can be expressed as \cite{hamilton2020graph}
\begin{equation}
    h_u^{(k+1)} = \psi\left(h_u^{(k)}, m_{\mathcal{N}(u)}^{(k)}\right) \label{eq:message passing}
\end{equation}
where $\psi$ is usually an MLP, and $m_{\mathcal{N}(u)}^{(k)}$ is the aggregated message
\begin{equation}
    m_{\mathcal{N}(u)}^{(k)} = \bigoplus_{v \in \mathcal{N}(u)}\left\{h_v^{(k)}, \forall v \in \mathcal{N}(u)\right\}
\end{equation}
for aggregation $\bigoplus_{v \in \mathcal{N}(u)}$, which must be permutation invariant \cite{hamilton2020graph}. Set pooling operations like sum ($\sum_{v \in \mathcal{N}(u)}$) or mean ($\frac{1}{N} \sum_{v \in \mathcal{N}(u)}$) are commonly used.

\begin{figure}[H]
    \centering
    \includegraphics[width=\linewidth]{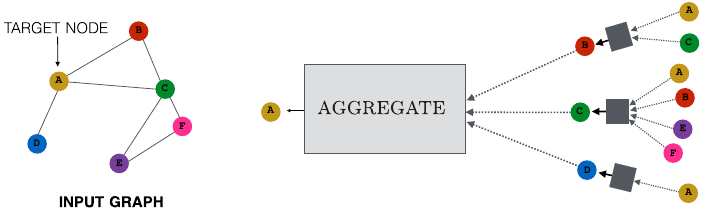}
    \caption{Overview of how a single node aggregates messages for a two-layer model. The GNN computation graph forms a tree structure as messages are sequentially aggregated from each node's local neighbourhoods. Figure taken from \cite{hamilton2020graph}.}
    \label{fig:GNN_message_passing}
\end{figure}

Depending on the type of GNN, multiple learnable MLPs can be included in the framework to increase expressivity \cite{zaheer_deep_2017, kipf_semi-supervised_2017}. However, we consider a simple architecture to simplify cumulant propagation. SAGEConv layers are implemented using \textsc{PyTorch Geometric} \cite{fey_fast_2019, hamilton_inductive_2017} with sum aggregation and a single linear transformation, and no biases or nonlinear activation functions. These layers follow the update rule
\begin{equation}
    h_u^{(k+1)} = W_{self}^{(k+1)} h_u^{(k)} + W_{neigh}^{(k+1)} \sum_{v \in \mathcal{N}(u)} h_v^{(k)}. \label{eq:graphsage update}
\end{equation}
Given input data $x \in \mathbb{R}^{N_0}$ we can iteratively apply this to obtain output node embeddings $y_u \in \mathbb{R}^{d_{out}}$.

\subsubsection{Decomposing GNNs as Mappings of Correlation Functions}
\label{subsec:GNN cumulants}
Here we extend the cumulant propagation framework in \prettyref{subsubsec:MLP cumulants} to GNNs. Similarly to MLPs, we consider layerwise cumulant propagation, and consider a generalised version of \prettyref{eq:graphsage update} with biases:
\begin{equation}
    z_u^{k} = W_{self}^k h_u^{k-1} + W_{neigh}^k \sum_{v \in \mathcal{N}(u)} h_v^{k-1} + b^k.
\end{equation}
Following \prettyref{eq:affine transf_re} we can write the cumulant generating function transformation as
\begin{align}
    \mathcal{W}_{z_u^k}(j) &= \ln \langle \exp(j^\top z_u^k) \rangle_{z_u^k} \nonumber \\
    &= \ln \Big\langle \exp \Big[ 
        j^\top W_{\text{self}}^k h_u^{k-1} \nonumber \\
    &\quad + j^\top W_{\text{neigh}}^k \sum_{v \in \mathcal{N}(u)} h_v^{k-1} 
        + j^\top b^k 
    \Big] \Big\rangle_{h_{u, v}^{k-1}}.
\end{align}
As the expectation is taken with respect to the joint distribution over all nodes within the neighbourhood of node $u_i$, this expression is difficult to evaluate. Guided by the CLT that uses i.i.d. variables, at a first approximation we can assume nodes are independent and write
\begin{align}
    \mathcal{W}_{z_u^k}(j) &= \mathcal{W}_{h_u^{k-1}}((W_{self}^k)^\top j) \nonumber \\
    &\quad + \sum_{v \in \mathcal{N}(u)} \mathcal{W}_{h_v^{k-1}}((W_{neigh}^k)^\top j) + j^\top b^k, \label{eq:GNN_cumulant_prop_general}
\end{align}
yielding for the first order cumulant ($n=1$)
\begin{align}
    G_{z_u^{k}}^{(1)} &= W_{self}^k G_{h_u^{k-1}}^{(1)} \nonumber \\
    &\quad + \sum_{v \in \mathcal{N}(u)} W_{neigh}^k G_{h_v^{k-1}}^{(1)} + b^k, \label{eq:GNN_cumulant_prop_mean}
\end{align}
and for $n \geq2$
\hspace*{-0.5em}
\begin{align}
    G_{z_u^{k},i_1,\dots,i_n}^{(n)} &= \sum_{s_1,\dots,s_n} W_{self, \;i_1 s_1}^k \cdots W^k_{self, \;i_n s_n} \, G_{h_u^{k-1},s_1,\dots,s_n}^{(n)} \nonumber \\
    &\hspace{-2.5em} + \sum_{\substack{v \in \mathcal{N}(u) \\s_1,\dots,s_n}} W_{i_1 s_1}^{k, neigh} \cdots W_{i_n s_n}^{k, neigh} G_{h_v^{k-1},s_1, \dots,s_n}^{(n)}. \label{eq:GNN_cumulant_prop_higher}
\end{align}
If a nonlinear activation function $\phi(z_u^k)$ were used, perturbative techniques could be used to determine the interactions between cumulants of different orders $n$. However, for simplicity we only consider this simple case of affine linear transformations.

\subsection{CLT: Analytical Solutions for Constrained MLP Weights}
\label{subsec:analytical weights}

Integrating inductive priors into network design can improve network training and generalisation to unseen tasks \cite{mccoy_modeling_2023, battaglia_relational_2018}. This section develops analytical solutions for symmetrically-constrained weights in simple MLP networks that implement the CLT transformation (\prettyref{subsec:RG_CLT}); their empirical generalisation is investigated in later sections.

\subsubsection{Linear Networks}

\prettyref{eq:laplace} shows how cumulants evolve under the CLT. We can therefore find equations relating input and output cumulants by using the propagation framework in \prettyref{subsubsec:MLP cumulants}. These equations can be inverted to solve for corresponding network weights. Results are given here, but for more detail see \prettyref{app:analytical solutions}.

\prettyref{eq:laplace} is interpreted as a decimation transformation from two to one dimensions, and it is clear that the transformation exhibits an $S_2$ symmetry corresponding to exchange of the variables $\xi_n^1$ and $\xi_n^2$. To `truly' learn this function, we expect learned network mappings should be invariant under the same symmetry, leading to constraints on the weights. In the simplest case of a network with a single linear layer and no biases, the weights are easily found to be
\begin{equation}
    W = \begin{pmatrix}
        \frac{1}{\sqrt{2}} & \frac{1}{\sqrt{2}}
    \end{pmatrix}. \label{eq:one layer weights}
\end{equation}

For the case of networks with no biases and a single two-dimensional hidden layer, we have the network mapping
\begin{align}
    y_i = g(x_i; \theta) = W_{ij}^2 W_{jk}^1 x_k
\end{align}
for $x = (\xi_n^1, \xi_n^2)^\top$. We seek $g(x) = g(Px)$ for permutation matrix
\begin{equation}
    P = \begin{pmatrix}
    0 & 1 \\
    1 & 0
    \end{pmatrix},
\end{equation}
leading to the constraint
\begin{equation}
    W^2 P^\top W^1 P = W^2 W^1.
\end{equation}
Solving this for $W^1$ we have
\begin{align}
    W^1 &= \begin{pmatrix}
        w_0 & w_1 \\ w_1 & w_0
    \end{pmatrix} \label{eq:symm W1} \\
    W^2 &= \begin{pmatrix}
        w_2 & w_2
    \end{pmatrix}, \label{eq:symm W2}
\end{align}
where $W^2$ has been chosen to treat both hidden units equally.

The CLT considers i.i.d. variables so we expect diagonal input cumulants, for example
\begin{align}
    G_x^{(1)} &= \begin{pmatrix}
        \kappa_1 \\ \kappa_1
    \end{pmatrix} \label{eq:G1}\\
    G_x^{(2)} &= \begin{pmatrix}
        \kappa_2 & 0 \\ 0 & \kappa_2
    \end{pmatrix} \label{eq:G2}.
\end{align}
\prettyref{eq:laplace} shows that cumulants obey a simple scaling law with $G_y^{(1)} = \sqrt{2} G_x^{(1)}, G_y^{(2)} = G_x^{(2)}, G_y^{(3)} = 1/\sqrt{2} G_x^{(3)}$, and so on. Transforming these first three cumulants according to \prettyref{eq:affine_mean_re} and \prettyref{eq:affine_higher_re} yields 3 conditions, which is enough to fully determine the weights in \prettyref{eq:symm W1} and \prettyref{eq:symm W2}. They form a surface in parameter space that satisfies
\begin{equation}
    w_0 = \frac{1}{w_2\sqrt{2}} - w_1. \label{eq:linear surface}
\end{equation}

We can introduce biases into these two-dimensional networks by similarly choosing them to treat both inputs equally; for the first layer we have $b = (b_1, b_1)^\top$, and as the output is one-dimensional we have $b_2$ only for the second layer. Extra parameters required more conditions from higher order cumulants. We again find that weights all lie in a subspace given by
\begin{align}
    w_0 &= \frac{1}{w_2 \sqrt{2}} - w_1 \nonumber \\
    b_2 &= -2w_2 b_1. \label{eq:subspace with biases}
\end{align}

\subsubsection{Quadratic Nonlinearity}
\label{subsubsec:quadratic nonlinearity}
We can use results given in \prettyref{app:statistical model} for cumulant propagation across a quadratic nonlinearity $\phi(z) = z + \alpha z^2$ to solve for network weights as above. Using the same network architecture, with no biases and the weights in \prettyref{eq:symm W1} and \prettyref{eq:symm W2}, we find that the first cumulant transforms as
\begin{align}
    \kappa_1 \sqrt{2} &= 2 w_2 \Big[ (w_0 + w_1)\kappa_1 + \alpha \kappa_1^2 (w_0 + w_1)^2 \nonumber \\
    &\quad + \alpha \kappa_2 (w_0^2 + w_1^2) \Big].
\end{align}
The nonlinearity mixes cumulants of different orders, yet \prettyref{eq:laplace} suggests a clean scaling of each $G_x^{(n)}$. As this expression should be true for any cumulants $\kappa_1, \kappa_2$, by comparing coefficients we find that
\begin{align}
    2 w_2 (w_0 + w_1) &= \sqrt{2} \nonumber \\
    2w_2 \alpha (w_0 + w_1)^2 &= 0 \nonumber \\
    2w_2 \alpha (w_0^2 + w_1^2) &= 0. \label{eq:inconsistency}
\end{align}
These equations have an inconsistency, suggesting that networks with a quadratic nonlinearity and symmetric weights cannot truly learn the transformation in \prettyref{eq:laplace}, and any learnt solution must be a poor approximation.

\section{Generalisation of MLPs}
\label{sec:MLP_gen}
\subsection{Tasks}
\label{subsec:MLP tasks}
In keeping with the probabilistic interpretation of the RG, we probe the generalisation of MLPs via two distinct tasks designed to test different distributional shifts. In both cases, the quality of learned transformations was assessed by networks' ability to successfully capture distributional structure of unseen i.i.d. inputs (see \prettyref{subsec:evaluation metrics}), which are all expected to tend to Gaussians under the CLT.

The \textit{Gaussian task} considers input Gaussians with varying variance over a single RG step (i.e., one iteration of the network), evaluating generalisation under shifts that preserve low-order cumulants, e.g. mean and variance. In contrast, the \textit{uniform task} evolves off-centre uniform distributions on the interval $[0,1]$ over multiple composed RG iterations, introducing non-zero higher-order cumulants with more complex behaviour. Both tasks were used to study the effect of symmetry and nonlinearity within the network by varying activation functions and weight constraints.

\subsection{Network Architecture and Training Setup}
\label{subsec:MLP training setup}

To directly investigate the effect of symmetry constraints discussed in \prettyref{subsec:analytical weights}, we used shallow networks with fixed width $N_l = 2$ and a single $l = 1$ hidden layer that follow the architecture defined in \prettyref{subsubsec:MLP architecture}. Networks were trained under explicit weight constraints as in \prettyref{eq:symm W1} and \prettyref{eq:symm W2}, and different activation functions were used to systematically vary both the degree of nonlinearity and extent of asymmetry of the activation function itself. For example, the quadratic activation $\phi(z) = z + \alpha z^2$ is minimally nonlinear yet preserves symmetry, while ReLU is both strongly nonlinear and asymmetric. This dual variation allows disentanglement of the impact of architectural and functional symmetry on generalisation.

In the \textit{Gaussian task} we tested purely linear networks, minimally nonlinear networks using the quadratic activation $\phi(z) = z + \alpha z^2$ with $\alpha = 0.5$, and strongly nonlinear networks with the ReLU activation \cite{nair_rectified_nodate}. The \textit{uniform task} additionally considered leaky ReLU activations \cite{xu_empirical_2015} with gradients in the $x  \leq 0$ region ranging from $1$ (effectively a linear network with no activation) to $0$ (standard ReLU), to interpolate between these extremal cases. To investigate how more expressive learnable functions might improve generalisation, a spline nonlinearity with trainable parameters was also considered both with and without frozen weights. Catmull-Rom-like cubic b-splines were used \cite{catmull_class_1974} with a smoothness regularisation term in the loss to discourage overfitting, and the interpolation range set adaptively for each network initialisation by considering the range of the spline inputs after the first affine layer.

All synthetic training and test data sets consisted of i.i.d. sample pairs drawn from the relevant distributions, with ground truth values comprising of their decimation outputs. For the uniform task, shuffled duplicates of network outputs were used as input variables in consecutive RG to remove all data correlations and ensure i.i.d. behaviour. Due to numerical sampling fluctuations, for each task a disparity was observed between empirical ground truth cumulants and the average values of those predicted from analytically scaling input cumulants as per \prettyref{eq:laplace}. Additionally, empirical input tensor cumulants were not exactly diagonal, as expected in e.g. \prettyref{eq:G2} due to finite sampling. A fixed data set size of $ n=10^6$ was used to balance cumulant estimation accuracy and computational speed (see \prettyref{app:exp setup} for more). Networks were implemented using \textsc{PyTorch} \cite{paszke_pytorch_2019}, and code is available \href{https://github.com/cass24687/Part-III-Project-Repo}{here}. Further key hyperparameters and training details are contained in \prettyref{app:exp setup}.

\subsection{Evaluation Metrics}
\label{subsec:evaluation metrics}

While standard generalisation metrics such as classification accuracy or raw MSE are commonly used in supervised learning benchmarks \cite{russakovsky_imagenet_2015, lecun_deep_2015}, in this context they are not appropriate. Rather than pointwise predictive performance, the focus is instead on evaluating the quality of learned transformations that preserve distributional structure. We therefore adopt empirical cumulant deviation and Kullback-Leibler (KL) divergence $D_{KL}$ as primary metrics to assess generalisation, which better quantify how well the transformed outputs preserve statistical alignment with the ground truth. Empirical cumulant deviation measures the difference between individual cumulants of the network output $p_{\text{net}}(y)$ and the ground truth distribution $p_{\text{gt}}(y)$. $D_{KL}$ provides a quantitative measure of the disagreement between two probability distributions \cite{kullback_information_1951}; here, we use it to more holistically assess how closely $p_{\text{net}}(y)$ matches $p_{\text{gt}}(y)$. Following Fischer et al., to account for any variability of output distributions across different network realisations, we normalise $D_{KL}$ by the Shannon entropy $H$ \cite{shannon_mathematical_1948} of the model output, yielding
\begin{equation}
    \hat{D}_{KL} (p_\text{net}\|p_\text{gt}) = \frac{D_{KL}(p_\text{net} \|p_\text{gt})}{H(p_\text{net})}.
\end{equation}

We compute averaged empirical cumulants up to fourth order from multiple repeat experiments. The variance of sample cumulants of order $n$ scales roughly as $\mathcal{O}(1/N)$ where $N$ is the number of repeats \cite{kendall_kendalls_1987}; five different seeds were used to suitably reduce estimation noise. As tensor input cumulants are required for propagation (\prettyref{subsubsec:MLP cumulants}), when investigating this framework higher order cumulants were explicitly symmetrised to preserve the expected property of multilinear symmetry \cite{mccullagh2018tensor}. Off-diagonal elements were retained to reflect sampling imperfections in their corresponding input distributions.

The discrete KL divergence was calculated between empirical histograms of network outputs and ground truth values. Where the cumulant values only were obtained,the KL divergence was determined with reconstructed probability densities using Fourier inversion of the characteristic function. Histogram bins were aligned to the reconstruction grid, and a frequency cutoff $t_\text{max} = 80$ and $2^{13}$ frequency sampling points were suitable to prevent aliasing while capturing the full range of the characteristic function.

\subsection{Gaussian Task Results}
\subsubsection{Linear Networks}
\begin{figure}[H]
  \centering
  \includegraphics[width=\linewidth]{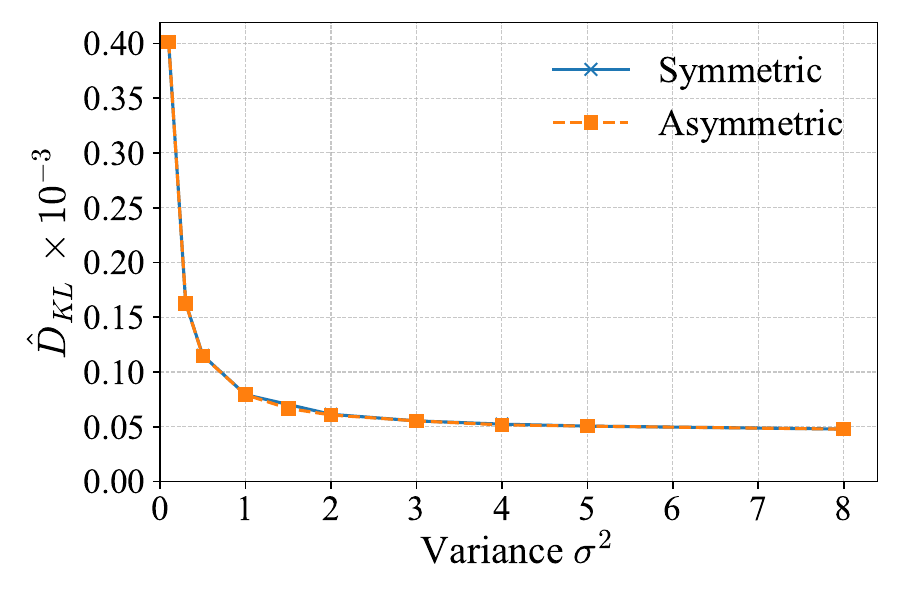}
  \caption{Normalised KL divergence $\hat{D}_{KL}(p_\text{net}\|p_\text{gt})$ as a function of input distribution variance $\sigma^2$ for linear networks. Error bars were negligible and are omitted.}
  \label{fig:linear_KL}
\end{figure}

We begin by evaluating the effect of symmetric weight constraints on the generalisation of linear networks. The normalised KL divergence between network output and ground truth distributions $\hat{D}_{KL}(p_\text{net}\|p_\text{gt})$  decreases inversely with input variance, suggesting networks improved generalisation at larger variances (\prettyref{fig:linear_KL}). In all cases $\hat{D}_{KL}$ remains very small, of order $10^{-4}$, with minimal variation between the two models. Additionally, the trained symmetric weights were found to satisfy the constraint equations \prettyref{eq:subspace with biases} to within a numerical tolerance of $10^{-4}$, validating the above analysis. These results establish a baseline for comparison with more expressive network architectures in later sections. 

\begin{figure*}[!t]
  \centering
  \includegraphics[width=\linewidth]{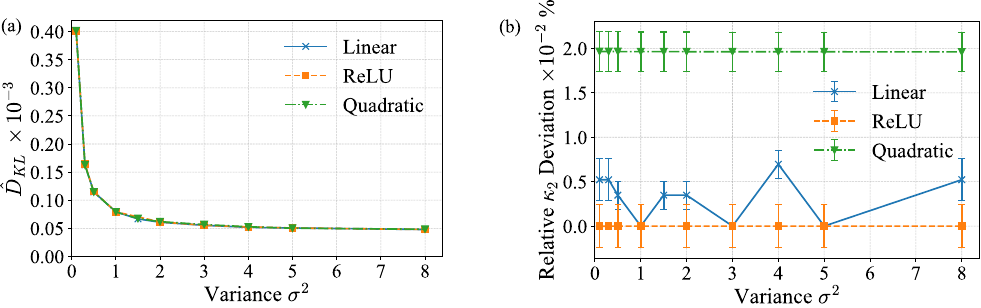}
  \caption{(a) Normalised KL divergence $\hat{D}_{KL}(p_\text{net}\|p_\text{gt})$ as a function of input variance $\sigma^2$. (b) Fractional deviations of second order cumulants $\kappa_2$ between network outputs and ground truth, expressed as percentages, as a function of variance $\sigma^2$. The three conditions shown all use unconstrained asymmetric weights. Error bars were negligible in (a) and are omitted.}
  \label{fig:nonlinear_asymm}
\end{figure*}

\begin{figure*}[!b]
  \centering
  \includegraphics[width=\linewidth]{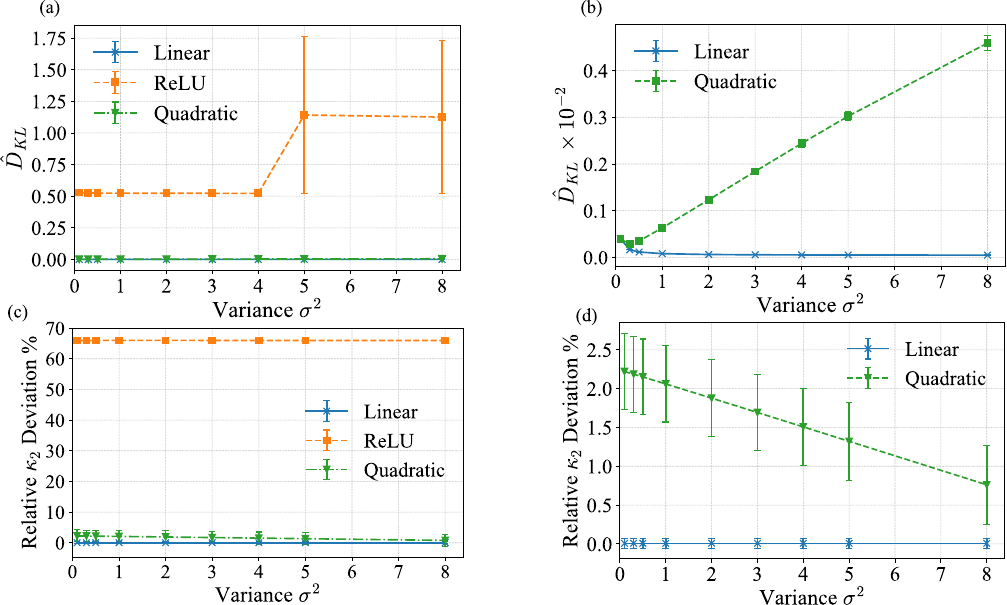}
  \caption{ (a), (b) Normalised KL divergence $\hat{D}_{KL}(p_\text{net}\|p_\text{gt})$ as a function of input variance $\sigma^2$. (c), (d) Relative deviations of second order cumulants $\kappa_2$, expressed as percentages, as a function of variance $\sigma^2$. Left hand panels show all three activation types (linear, ReLU, quadratic); right hand panels show only linear networks and quadratic activations. All networks use symmetrically-constrained weights.}
  \label{fig:nonlinear_symm}
\end{figure*}

\subsubsection{Nonlinear Networks with Asymmetric Weights}
\label{subsubsec:gaussian_nonlinear_asymm}
We next evaluate the impact of nonlinearity and asymmetry within the activation function on generalisation, using networks with unconstrained asymmetric weights. There is minimal variation in $\hat{D}_{KL}(p_\text{net}\|p_\text{gt})$ across linear, ReLU, and quadratic activations, with all models exhibiting improved generalisation at larger input variances (\prettyref{fig:nonlinear_asymm}). However, these similarities do not extend to the individual cumulants. While both linear networks and those with ReLU activation predict target $\kappa_2$ values well, networks with quadratic activation consistently overestimate $\kappa_2$ by approximately 0.02\% for all input variances. Although deviations are all small, these differences are systematic, highlighting the increased expressivity of stronger nonlinearities. Higher order cumulants ($n > 2$) vanish for Gaussian distributions so we present results for $\kappa_2$ only.

\subsubsection{Nonlinear Networks with Symmetric Weights}
\label{subsubsec:nonlinear symm}
We now consider the combined effect of applying symmetry constraints to both network weights and activation functions. The results in \prettyref{fig:nonlinear_symm} show a marked difference in behaviour from the unconstrained case in \prettyref{fig:nonlinear_asymm}. $\hat{D}_{KL}$ for the ReLU activation exhibits a dramatic increase, being approximately four orders of magnitude greater, and for the quadratic activation $\hat{D}_{KL}$ similarly increases by a factor of ten. These discrepancies are reflected in the $\kappa_2$ deviations. The ReLU activation displays a very high $\kappa_2$ deviation that systematically exceeds 60\%, while the deviation for the quadratic nonlinearity is two orders of magnitude greater than in the asymmetric case. 

The expected trend of $\kappa_2$ deviation decreasing with variance is observed for linear and quadratic networks. In contrast, $\hat{D}_{KL}$ for both ReLU and quadratic activations increases with variance. This behaviour suggests poor matching of higher order cumulants, highlighting the brittleness of the learned transformations. These results are consistent with those of \prettyref{subsubsec:quadratic nonlinearity}, where it is shown that no analytical solution exists for networks of this size with symmetric weights and quadratic nonlinearities. The findings here indicate that strict symmetry constraints can hinder network expressivity when symmetry breaking is required for learning.

\subsection{Uniform Task Results}

Linear networks exhibited similar generalisation performance as in the Gaussian task, with minimal difference between networks with symmetric and unconstrained weights and small $\hat{D}_{KL}$ values. We therefore do not focus explicitly on their behaviour, instead using them as a benchmark for comparison where appropriate. Additionally, as the uniform task is designed to assess generalisation under transformations involving higher order cumulants, we present results for $\kappa_4$ only, with those for $\kappa_2$ and $\kappa_3$ provided in certain cases in \prettyref{app:extra MLP results}.

\subsubsection{Quadratic Activation}

This section continues the discussion of task-dependent symmetry breaking by focusing on networks with quadratic nonlinearity and examining the effects of symmetric and unconstrained weights on generalisation. The results in \prettyref{fig:uniform_quadratic} are concordant with both \prettyref{subsubsec:nonlinear symm} and \prettyref{subsubsec:quadratic nonlinearity}: networks with unconstrained weights significantly outperform their symmetric counterparts, which quickly accumulate error over successive RG steps. In the symmetric case, $\hat{D}_{KL}$ increases by up to four orders of magnitude compared to the linear baseline, while networks with asymmetric weights maintain values close to zero throughout. Similar behaviour is observed for the relative $\kappa_4$ deviation, which not only diverges to approximately $60\times10^3 \%$ in the symmetric case, but changes sign too, highlighting the breakdown in the learned transformation's ability to preserve higher order distributional structure.

\subsubsection{Leaky ReLU with Asymmetric Weights}
\label{subsubsec:leaky_relu_asymm}
\begin{figure*}[!t]
  \centering
  \includegraphics[width=\linewidth]{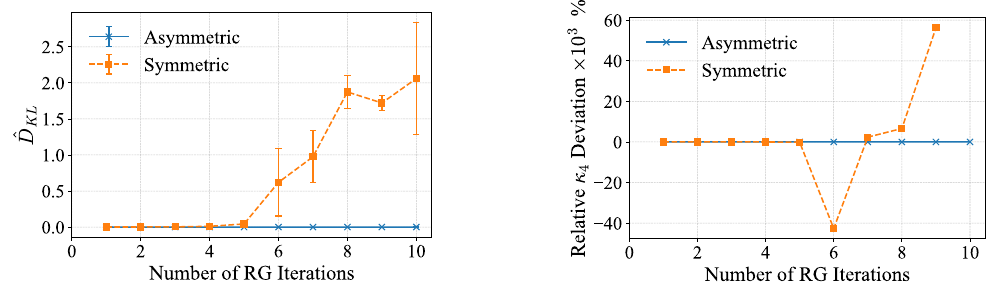}
  \caption{ (a) Normalised KL divergence $\hat{D}_{KL}(p_\text{net}\|p_\text{gt})$, and (b) relative $\kappa_4$ deviation, both plotted as a function of the number of successive RG steps. Both panels show networks with constrained and unconstrained weights.}
  \label{fig:uniform_quadratic}
\end{figure*}

\begin{figure*}[!b]
  \centering
  \includegraphics[width=\linewidth]{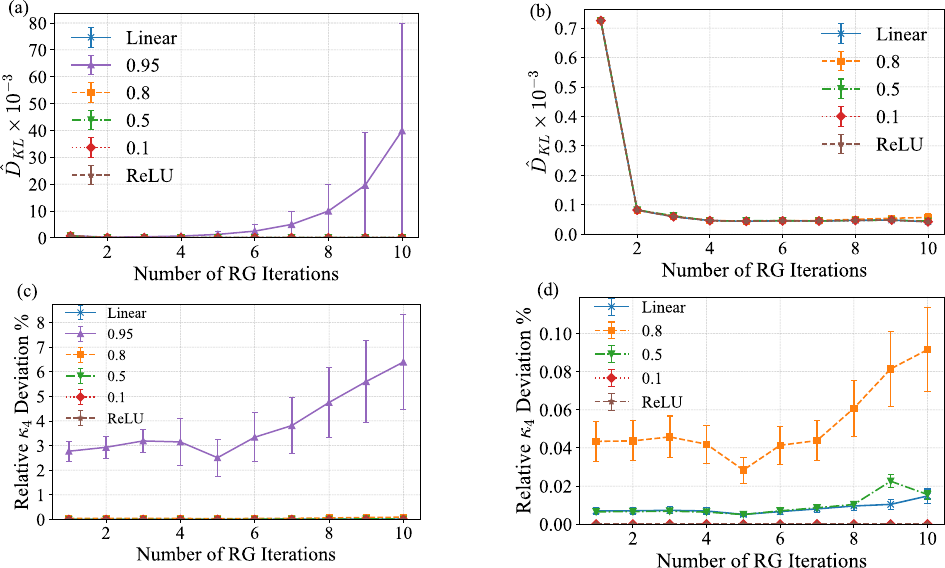}
  \caption{ (a), (b) Normalised KL divergence $\hat{D}_{KL}(p_\text{net}\|p_\text{gt})$, and (c), (d) relative $\kappa_4$ deviation, both as a function of the number of RG steps. Left and right hand panels are plotted with and without $0.95$ gradient. All networks have unconstrained weights.}
  \label{fig:asymm_leakyrelu_main}
\end{figure*}

Here we consider networks with asymmetric weights and leaky ReLU activation, and smoothly vary the negative slope from $1$ (a linear network) to $0$ (corresponding to standard ReLU) to isolate the role that asymmetry and nonlinearity within the activation function plays in learning. \prettyref{fig:asymm_leakyrelu_main} shows that a gradient of $0.95$ exhibits particularly poor generalisation. For this case, $\hat{D}_{KL}$ reaches up to $0.04$; while significantly smaller than those observed for the quadratic nonlinearity in \prettyref{fig:uniform_quadratic}, values are still two orders of magnitude larger than the linear benchmark. There is little variation in $\hat{D}_{KL}$ amongst the other gradients considered, though a consistent spike is observed across all conditions at the first RG step.

 \begin{figure}[H]
  \centering
  \includegraphics[width=\linewidth]{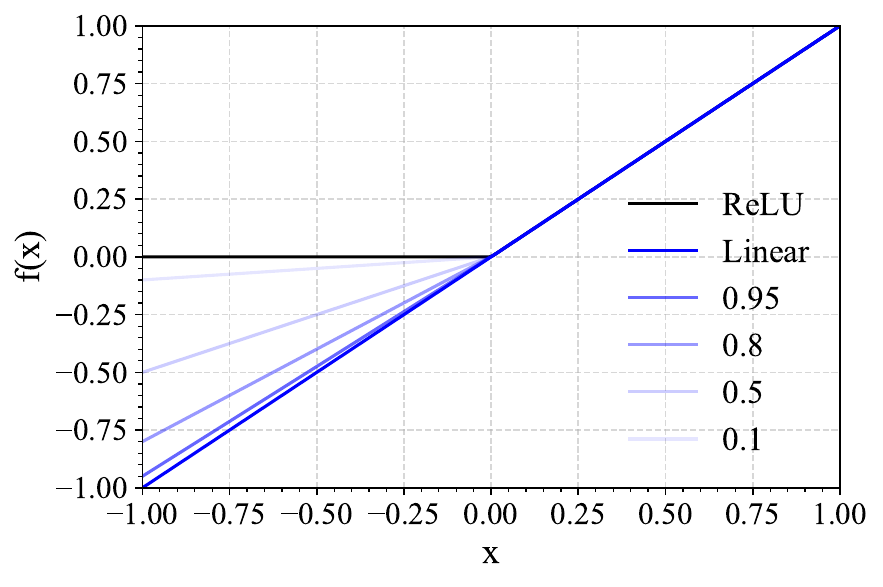}
  \caption{Schematic showing the different slopes of leaky ReLU investigated.}
  \label{fig:leaky_slopes}
\end{figure}

The $\kappa_4$ deviation plots reveal a more nuanced picture. Despite the fact that values are again substantially smaller than for the quadratic nonlinearity, there is a clear performance hierarchy. Networks with slope $0.5$ perform comparably to purely linear equivalents, with deviations of approximately $0.01\%$. However, slightly perturbing the gradient to $0.95$ drastically reduces performance, which improves as the degree of nonlinearity is increased, until networks outperform the linear baseline. This behaviour resembles a symmetry-breaking transition, analogous to phase transitions in physical systems like ferromagnets, where a small change in some control parameter induces a sharp qualitative shift in the system. Similar trends are observed in $\kappa_2$ and $\kappa_3$ and are included in \prettyref{app:extra MLP results}.

\subsubsection{Leaky ReLU with Symmetric Weights}

\begin{figure*}[!t]
  \centering
  \includegraphics[width=\linewidth]{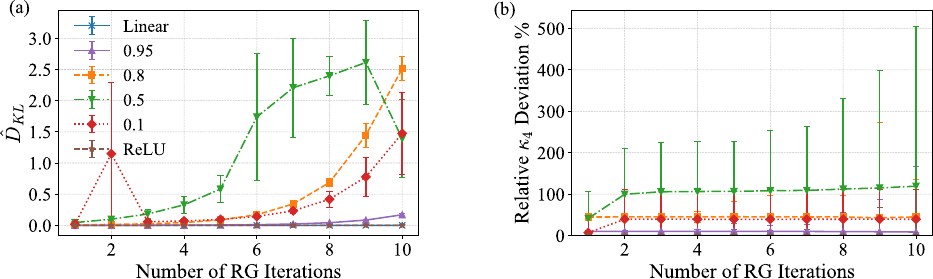}
  \caption{ (a) $\hat{D}_{KL}(p_\text{net}\|p_\text{gt})$, and (b) relative $\kappa_4$ deviation, both as a function of RG step. Both panels are plotted with the $0.95$ gradient condition. All networks have symmetric weights.}
  \label{fig:symm_leakyrelu_main}
\end{figure*}

In this section the same analysis as \prettyref{subsubsec:leaky_relu_asymm} is repeated but using symmetric weights. \prettyref{fig:symm_leakyrelu_main} shows that performance is generally worse across the different leaky ReLU gradients than in the previous asymmetric weights case. The $\hat{D}_{KL}$ values are two orders of magnitude greater, with larger errors that suggest increased variability across network initialisations. This behaviour is reflected in $\kappa_4$ deviation values, which are three orders of magnitude greater than the asymmetric case. Both metrics have large errors that stem from reduced training stability.

Both the linear and ReLU conditions generalise the best, and exactly the same as each other, while the $0.5$ gradient is clearly the worst. At these intermediate slopes, neither symmetric nor asymmetric structure dominates, reducing the network's ability to generalise. This behaviour is reminiscent of frustrated regimes in physical systems, where competing symmetries or constraints prevent optimal ordering. Furthermore, the relative $\kappa_4$ deviation plateaus after just two RG iterations and remains constant thereafter, while the $\hat{D}_{KL}$ continues to grow. This difference highlights the architectural bottleneck: the network rapidly converges to a limited internal representation which causes higher order cumulants to decay at a rate that is offset by some fixed amount to the true transformation, and the full distributional mismatch increases. Results for $\kappa_2$ and $\kappa_3$ are included in \prettyref{app:extra MLP results}.

\begin{figure}[H]
  \centering
  \includegraphics[width=0.95\columnwidth]{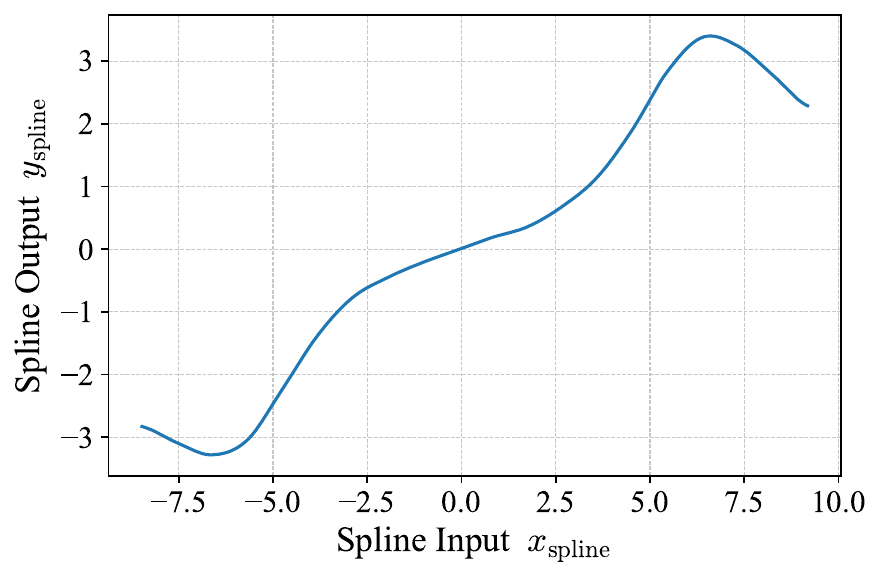}
  \caption{Unstable spline shape when only spline parameters are trainable.}
  \label{fig:only_unfrozen_spline}
\end{figure}

\begin{figure}[h]
  \centering
  \includegraphics[width=2.2\linewidth]{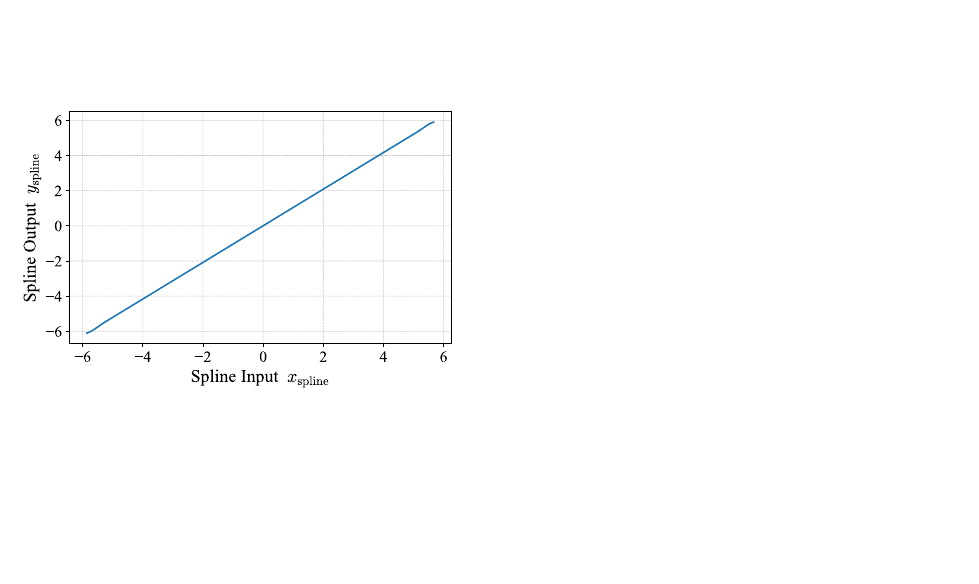}
  \caption{Linear spline shape when both weights and spline parameters are learnable.}
  \label{fig:both_unfrozen_spline}
\end{figure}

\subsubsection{Spline Activation with Asymmetric Weights}

In this section, we consider networks with a learnable spline nonlinearity under two conditions of fixed or trainable weights. We did not impose symmetry constraints on weights in either case. Rather than directly investigating symmetry preservation of the transformation, the focus was instead on how generalisation is affected when models are allowed to share the learning load between the activation function and weights.

\begin{figure*}[!b]
  \centering
  \includegraphics[width=\linewidth]{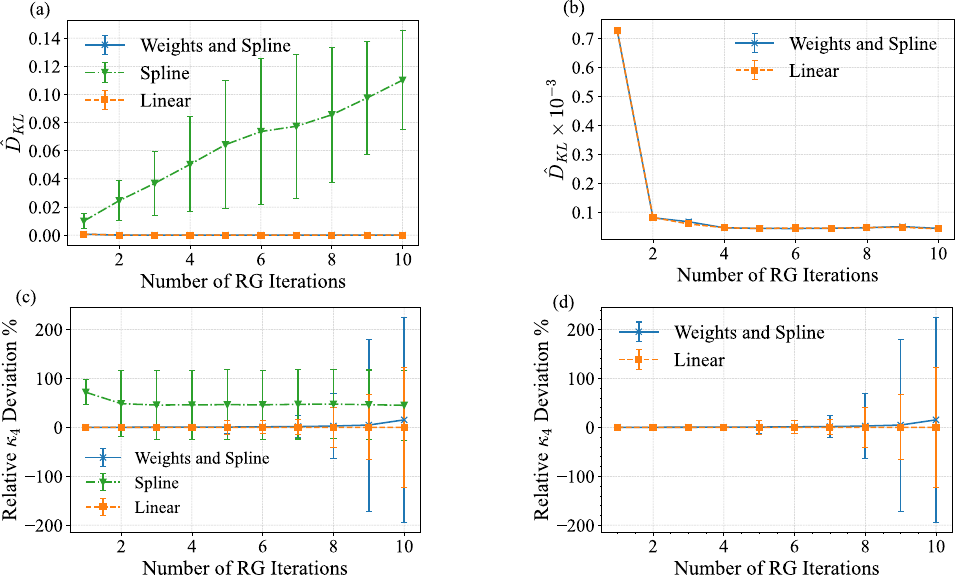}
  \caption{ (a), (b) $\hat{D}_{KL}(p_\text{net}\|p_\text{gt})$, and (c), (d) relative $\kappa_4$ deviation, both as a function of RG step. Legend labels refer to what parameters are trainable, e.g. `Spline' corresponds to only having unfrozen spline parameters. Right hand panels are plotted without this `Spline' condition.}
  \label{fig:spline_main}
\end{figure*}

As shown in \prettyref{fig:spline_main}, the `Spline' configuration with only learnable spline parameters did not generalise well. It quickly accumulated large errors, with $\hat{D}_{KL}$ values increasing roughly tenfold over the total number of iterations and the relative $\kappa_4$ deviation constant at approximately $50\%$. Expressivity is severely limited by fixed weights that vary randomly with initialisation, which the spline compensates for with unstable and overfitted spline shapes as in \prettyref{fig:only_unfrozen_spline}.

In contrast, when both the weights and spline parameters are trainable, networks learn a linear architecture with spline shapes remaining linear as in \prettyref{fig:both_unfrozen_spline}. Both the $\hat{D}_{KL}$ and $\kappa_4$ deviations match almost perfectly with the linear baseline, and values are very small for both. These results show that network expressivity should be guided by the task at hand; when given the freedom to do so the model prefers to minimise unnecessary nonlinearity, with excessively flexible activations potentially degrading performance.

It should be noted that the fact that $\hat{D}_{KL}$ values remain small while the $\kappa_4$ deviation diverges at high RG iterations suggests that this behaviour is likely an insignificant numerical artefact. At higher iterations $\kappa_4$ shrinks as the distribution becomes increasingly Gaussian, so very small cumulant values can manifest as large relative errors.

\section{Generalisation of GNNs}
\label{sec:GNN_gen}

\subsection{Tasks}

In this section we investigate the effect of architectural inductive biases introduced by GNNs on the generalisation of learned CLT transformations. We consider the same uniform task described in \prettyref{subsec:MLP tasks}, involving distributional shifts with higher-order cumulants over multiple composed RG steps. Generalisation is assessed via empirical cumulant deviation and KL divergence as before, averaging measurements over five different seeds to reduce estimation noise. When only cumulant information was available, the same Fourier inversion of the characteristic function was applied to reconstruct probability density functions, with the same frequency resolution and cutoff. 

GNNs are typically applied to large symmetric graphs where their built-in global permutation equivariance plays a significant role \cite{xu_how_2019}. However, rather than comparing performance amongst GNNs, the aim here is instead to better understand how their local aggregation mechanisms and architectural biases influence generalisation relative to MLPs. To this end, we apply GNNs to simple two-node directed graphs that mirror the two-to-one dimensionality reduction of \prettyref{eq:laplace}, avoiding complexity from larger irregular graphs. Although these graphs lack node-level permutation symmetry, their use still enables comparison between the message-passing formalism and feedforward MLP architectures.

\subsection{Network Architecture and Training Setup}
\label{subsec:GNN architecture}

We use shallow GNN architectures with either one or two layers, and low-dimensional hidden representations (1D or 2D), to minimise model capacity and reduce the risk of overfitting. This is appropriate given the small input graph size and scalar node features. Models were all implemented using \textsc{PyTorch Geometric} \cite{fey_fast_2019}. To simplify analysis and support verification of the extended GNN cumulant propagation framework, we use linear SAGEConv layers \cite{hamilton_inductive_2017} without any activation functions. Further hyperparameters and training details are included in \prettyref{app:exp setup}.

\subsection{Cumulant Propagation Validation}

Here we evaluate the validity of the cumulant propagation framework under the independent node approximation that is encapsulated in \prettyref{eq:GNN_cumulant_prop_general}. We consider the uniform task, and cumulants are propagated across networks layers within each step by iteratively applying \prettyref{eq:GNN_cumulant_prop_mean} and \prettyref{eq:GNN_cumulant_prop_higher} to the two-node directed graph we consider. For comparison, we use a linear MLP baseline with asymmetric weights and two affine layers, with cumulant propagation performed using \prettyref{eq:affine_mean_re} and \prettyref{eq:affine_higher_re}.

\begin{figure*}[!t]
  \centering
  \includegraphics[width=\linewidth]{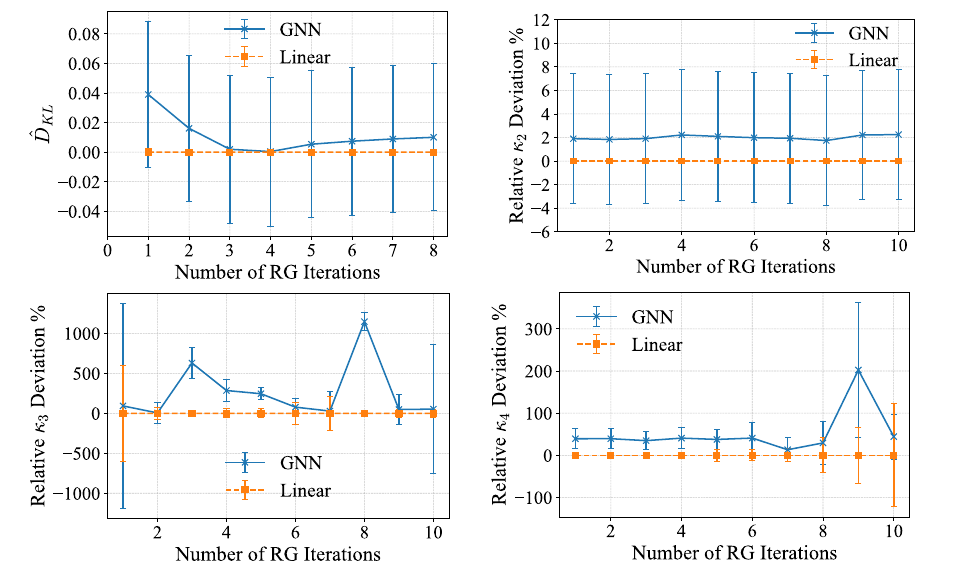}
  \caption{$\hat{D}_{KL}(p_\text{prop} \|p_\text{gt})$ and relative cumulant deviations for all of $\kappa_2$, $\kappa_3$, $\kappa_4$. Propagation breaks down for higher order cumulants. While not shown, $\hat{D}_{KL}$ values increase rapidly above 8 RG iterations; the plot instead focuses on the behaviour relative to the linear baseline for lower RG steps.}
  \label{fig:gnn_propagation}
\end{figure*}

As shown in \prettyref{fig:gnn_propagation}, the GNN is able to match the MLP baseline well to an extent, with initially small $\hat{D}_{KL}(p_\text{prop} \|p_\text{gt})$ values of order $10^{-2}$ and relative $\kappa_2$ deviations consistently within $2\%$. However, the framework is clearly limited. The $\hat{D}_{KL}$ and $\kappa_2$ deviation plots show much larger errors for the GNN than the MLP baseline, and the GNN fails to capture higher order cumulants, with relative deviations in $\kappa_3$ and $\kappa_4$ exceeding several hundred percent. These limitations are attributed to the crudeness of the independent node approximation. While sufficing for low order cumulants, it neglects important interactions induced by message passing. While node embeddings are initially independent from i.i.d. sampling, during training node representations become correlated both through the aggregation of neighbour information and the fact that weights are globally shared across nodes. These dependencies are essential for network learning dynamics and are not reflected in our framework, underlying its reduced performance.

\begin{figure*}[!t]
  \centering
  \includegraphics[width=0.9\linewidth]{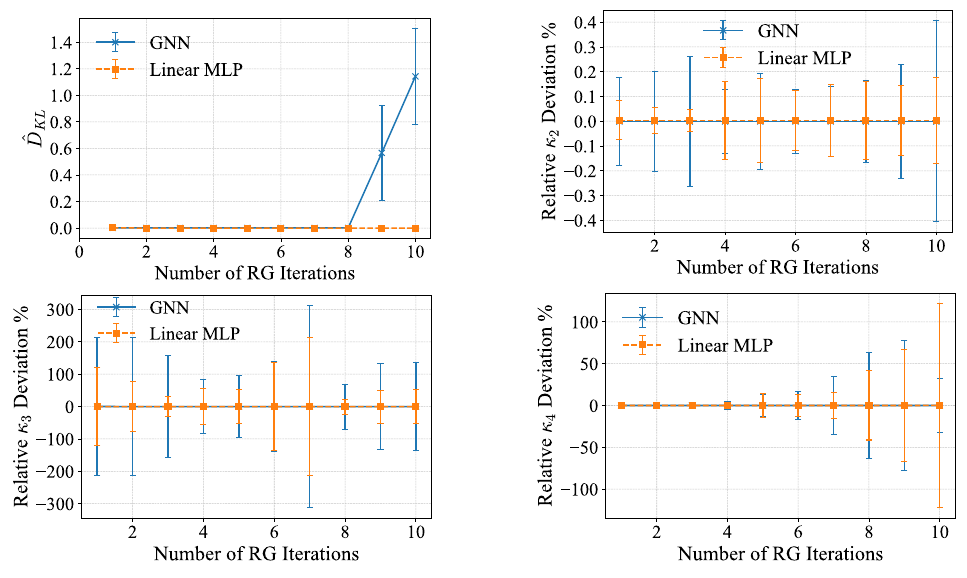}
  \caption{$\hat{D}_{KL}(p_\text{net} \|p_\text{gt})$ and relative cumulant deviations for all of $\kappa_2$, $\kappa_3$, $\kappa_4$. Generalisation is at most as good as the linear baseline.}
  \label{fig:gnn_generalisation}
\end{figure*}

\subsection{Uniform Task Generalisation Results}

The GNN architecture considered performs comparably to the MLP baseline (\prettyref{fig:gnn_generalisation}). $\hat{D}_{KL}(p_\text{net} \|p_\text{gt})$ values are consistently very small, approximately of order $10^{-4}$, and the relative cumulant deviations match well. However, the GNN does show some limitations. Despite being initially of small magnitude, the GNN $\hat{D}_{KL}$ values quickly spike above 8 RG steps, and the errors in relative GNN cumulant deviations are generally at least as large as those of the MLP. 

Although GNNs' inductive biases can be very useful for arbitrarily complex structured data, these same biases can be counterproductive when the task lacks the structure that the GNN is designed to exploit. Here, the two-node directed graph has minimal symmetry or topological complexity, and does not present any permutation structure for the GNN's permutation-equivariant operations to exploit. As a result, MLPs generalise slightly more effectively, with the GNN architecture possibly hindering learning by imposing inductive biases misaligned with the task.

Finally, we note that both models superficially appear to consistently struggle predicting $\kappa_3$, and fail to capture $\kappa_4$ well at high RG iterations. However, these are most likely numerical artefacts: for the symmetric uniform distribution $\kappa_3$ is $0$, and both cumulants decay under the CLT, resulting in large relative errors.


\section{Discussion}
\label{sec:discussion}

This work primarily explores the role of symmetry and architectural expressivity in improving the generalisation of learned decimation RG transformations. Symmetry constraints on network parameters can often lead to improved parameter efficiency and better generalisation \cite{hu_exploring_2019, nachum_symmetry_2020}. However, we present a more nuanced perspective: networks must preserve the symmetries encoded in the RG transformation, but there is a complex interplay between symmetry constraints and network expressivity capacity, and symmetry constraints only aid generalisation when aligned with the representational demands of the task. This closely parallels the work of Ziyin et al. \cite{ziyin_parameter_2025}, who argue that the principle of symmetry breaking and restoration in network parameters can unify theories of deep learning.

We observe that in linear networks, symmetry has little effect due to limited expressivity. For networks using the quadratic activation, symmetric weights overconstrain the system and generalisation is poor. We show analytically that no solution exists for the small MLPs considered. ReLU activations offer stronger nonlinearity and perform even worse in the Gaussian task. In these cases, symmetry breaking is required for learning. Just as Goldstone modes emerge when continuous symmetries are broken in physical systems \cite{peskin2018introduction}, asymmetric weights provide more representational degrees of freedom that are necessary for effective learning. These results are consistent with prior works that indicate how networks break symmetry to adapt to different tasks \cite{fok_spontaneous_2017}.

In leaky ReLU networks with unconstrained weights we observe a sharp phase transition-like phenomenon. Models with weak nonlinearities show considerably worse generalisation than linear equivalents, but performance improves as nonlinearity increases. In contrast, symmetric weights clash with the internal asymmetry of the activation, resulting in frustrated learning dynamics. This competition between symmetry and expressivity echoes ideas from condensed matter physics, where frustrated systems cannot simultaneously minimise all local energy constraints.

Our work with both spline activations and GNNs reinforces this narrative. In both cases, the architectural inductive biases acted as an Occam's razor: minimally but sufficiently expressive systems generalised best. Our results suggest that network expressivity and architectural bias should be matched to the structure of the task. In the spline case, the needlessly complex activation with fixed weights was more prone to overfitting and poor generalisation. When the weights were trainable, the model effectively reduced to a linear MLP with improved performance, reminiscent of the double descent phenomenon in machine learning. For GNNs, the built-in permutation equivariance combined with the single directed edge was misaligned with the structure of the CLT, limiting generalisation. Such findings are consistent with behaviour observed in many modern neural models, for instance in shortcut learning \cite{geirhos_shortcut_2020}.

From a physics perspective, our results aim to aid understanding of how renormalisation concepts and associated symmetries can be encoded within neural models to improve generalisation. We build on established connections between RG and hierarchical representations in deep networks to gain further insight into their capacity to act as models of coarse-graining.

To interpret the internal information processing performed by GNNs, we also extend an existing framework that propagates cumulants through MLP models \cite{fischer_decomposing_2022} to this more complex architecture. We validate a simple first approximation that is able to consistently track the covariance over multiple RG steps to within $2\%$. However, it fails to capture crucial node correlations induced through training and could be improved upon with further theoretical modelling.

\vspace{1em}
There are several limitations to this study. Firstly, the low dimensionality of the CLT transformation considered may limit the generality of our results. Additionally, this transformation is a particularly simple case and future work could extend similar analyses to more complex RG flows. Coarse-graining in physical systems like the Ising model typically involve structured dependencies that can be analysed probabilistically using Dyson's hierarchical models \cite{bleher_investigation_1973, jona-lasinio_renormalization_2001} and offer a more realistic physics benchmark. GNNs are a natural architecture to capture such spatial dependencies. While prior work has evaluated the effect of equivariance on GNN generalisation \cite{maron_invariant_2019, keriven_universal_2019}, the full potential of their inductive biases is not fully harnessed by the simplified tasks considered here.

Additionally, the technique of analytically inverting cumulant propagation equations might not scale to more expressive or deeper models. Alternate approaches may be required to obtain similar analytical results to retain interpretability.

\section{Conclusions}
This work demonstrates the existence of a critical balance between symmetry constraints and network expressivity that determines generalisation performance. Focusing on the central limit theorem as a controlled test case of a renormalisation flow, our analytical framework, which extends cumulant propagation theory from MLPs to GNNs, provides a clear lens through which to interpret how network models learn physically meaningful transformations. Our results indicate that symmetry constraints must be applied carefully, with the potential to harm generalisation by being overly restrictive. Equally, more flexible architectures can lead to reduced performance through overfitting or poor inductive alignment, reinforcing the importance of matching architectural bias with specific tasks.

These observations suggest a broader design principle for neural network design in physics applications: effective architectures must leverage physical symmetries carefully, without constraining representational flexibility that is useful to capture essential features of underlying transformations. Besides investigation of more complex RG flows, future work could explore adaptively incorporating symmetry constraints during training (such as \cite{veefkind_probabilistic_2025}), which potentially offers a more nuanced approach to encoding physical priors in network models.


\begin{appendices}
    
\section{Central Limit Theorem Cumulant Recursion Relation}
\label{app:CLT recursion}
We start here from the one-step renormalisation relation between distributions, \prettyref{eq:distrib_recursion}:
\begin{equation}
    p_{n+1}(x) = \sqrt{2} \int dy\, p_n(\sqrt{2}x - y)\, p_n(y),
\end{equation}
and consider the moment generating function
\begin{equation}
    M_{n+1}(s) = \int_{-\infty}^{\infty} e^{s x} \, p_{n+1}(x) \, dx.
\end{equation}
Making the substitution $u = x \sqrt{2} - y$ with $dx = \frac{1}{\sqrt{2}}du$, the convolution becomes
\begin{equation}
    M_{n+1}(s) = \int du \int \, dy \, e^{\frac{s(u + y)}{\sqrt{2}}} \, p_n(u) \, p_n(y)
\end{equation}
which is separable, yielding
\begin{equation}
    M_{n+1}(s) = \left[ M_n\left( \frac{s}{\sqrt{2}} \right) \right]^2.
\end{equation}
The cumulant generating function is defined as $K_{n+1}(s) = \ln M_{n+1}(s)$ and the cumulants $\kappa_n^{(n)}$ are the derivatives
\begin{equation}
    \kappa_r^{(n)} = \left. \frac{d^r}{ds^r} K_n(s) \right|_{s=0}. 
\end{equation}
By applying the chain rule to
\begin{equation}
K_{n+1}(s) = \ln \left[ M_n\left( \frac{s}{\sqrt{2}} \right)^2 \right]
= 2 K_n\left( \frac{s}{\sqrt{2}} \right),
\end{equation}
and noting that the zeroth order cumulant is always zero from normalisation, we have the desired relation
\begin{equation}
    \kappa_r^{(n+1)} = 2^{1 - r/2} \, \kappa_r^{(n)} \quad \text{for } r \geq 1 \label{eq:desired}.
\end{equation}

This result suggests the following three conservation laws are associated with the transformation $\mathcal{R}$: normalisation, centreing, and variance. We have
\begin{align}
    \int p_{n+1}(x)\, dx &= \int p_n(x)\, dx , \\
    \int x p_{n+1}(x)\, dx &= \int x p_n(x)\, dx \label{eq:centreing}, \\
    \int x^2 p_{n+1}(x)\, dx &= \int x^2 p_n(x)\, dx \label{eq:cons_var}.
\end{align}
 A centred distribution therefore preserves its variance with all higher order cumulants vanishing, which is the essence of the CLT.




\section{MLPs as Mappings of Correlation Functions}
\label{app:statistical model}




This appendix contains more detail on the cumulant propagation framework outlined in \prettyref{subsubsec:MLP cumulants} for MLPs.

We first consider cumulant propagation across the nonlinear activation function $\phi$ by writing
\begin{align}
    \mathcal{W}_{y^{l}}(j) &= \ln \langle \exp(j^\top y^{l}) \rangle_{y^{l}} \nonumber \\
&= \ln \langle \exp(j^\top \phi(z^{l})) \rangle_{z^{l}}. 
\end{align}
As noted in the main text, this generating function cannot be evaluated exactly \cite{fischer_decomposing_2022} in general. Perturbative techniques that exist to overcome this difficulty proceed by replacing $\phi (z^{l})$ with its Taylor expansion $\sum_{m} \frac{\phi^{(m)}\big|_{z^l = 0}}{m!} (z^l)^m$, and treat the nonlinear terms ($m > 1$) as being small \cite{helias_statistical_2020}. By expanding the activation function in this way it is possible to determine analytic expressions for the cumulants $G_{y^{l}}^{(n)}$ of the postactivations $y^{l}$ in terms of those of the preactivations $z^l$ for arbitrary activation functions. More information is contained in \cite{helias_statistical_2020}; for example, how to construct postactivation cumulants as a series of Feynman diagrams. Cumulants $G_{y^{l}}^{(n)}$ of order $n$ are typically functions of multiple cumulants of the preactivations $z^{l}$ of different orders $m$, $G_{y^{l}}^{(n)} = f_{n}(\{ G_{z^{l}}^{(m)} \})$, representing the information transfer between layers. This behaviour is distinct from the affine transformations which do not mix cumulants. Such a  perturbative approach assumes a differentiable activation function, but it is possible to adapt the method for nondifferentiable functions such as ReLU by using a Gram-Charlier expansion of the probability distribution $p(z^{l})$ \cite{fischer_decomposing_2022}.

Our analytical results for symmetrically-constrained weights in MLPs hinge upon this propagation framework for a quadratic nonlinearity $\phi(z) = z + \alpha z^2$. This particular activation has relatively simple expressions for the interaction functions $f_{1, 2}(G_{z^{l}}^{(m)})$ of the first two cumulants $n = 1, 2$, the mean and variance. These are
\begin{subequations}
\begin{align}
    \mu_{y^l, i} &= \mu_{z^l, i} + \alpha (\mu_{z^l, i})^2 + \alpha \, \Sigma_{z^l, ii}\;, \label{eq:mean interaction function} \\[1.5ex] 
    \Sigma_{y^l, ij} &= \Sigma_{z^l, ij} + 2\alpha \, \Sigma_{z^l, ij} (\mu_{z^l, i} + \mu_{z^l, j}) \nonumber \\
    &\quad + 2\alpha^2 (\Sigma_{z^l, ij})^2 + 4\alpha^2 \mu_{z^l, i} \, \Sigma_{z^l, ij} \, \mu_{z^l, j} \nonumber \\
    &\quad + \alpha (1 + 2\alpha \, \mu_{z^l, i}) \, G_{z^l, (i,i,j)}^{(3)} \nonumber \\
    &\quad + \alpha (1 + 2\alpha \, \mu_{z^l, j}) \, G_{z^l, (j,j,i)}^{(3)} \nonumber \\
    &\quad + \alpha^2 \, G_{z^l, (i,i,j,j)}^{(4)}.
\end{align}
\end{subequations}

By composing the transformations for cumulants across affine layers, and the relevant interaction functions $f_{n}(\{ G^{(m)} \})$ for the choice of activation function, for each order $n$ we can write
\begin{align}
G_y^{(n)} &= W^{L+1} \left( f_{n} \left( \cdots \{ G_{x}^{(m)} \}_{m}  \cdots\right) \right) + b^{L+1} \nonumber\\
&=: g_{n} \left( \{ G_{x}^{(m)} \}_{m}; \theta, \phi \right).
\end{align}
As the nonlinearity mixes orders of cumulants, each cumulant of the network output will in general depend on the full set of input cumulants. The \textit{statistical model} of the network is then defined as the mapping
\begin{equation}
    g_{\text{stat}} : \left( \left\{ G_{x}^{(m)} \right\}_{m}, \theta, \phi \right) \mapsto p(y)
\end{equation}
where $p(y)$ is characterised by the propagated cumulants. It is important to note that there is a one-to-one correspondence between the statistical model and the network model $g\; :\; (x, \theta) \mapsto y$, as they both share the same parameters $\{ W^l, b^l \}_{l=1,\dots,L+1}$.

\vspace{1em}
Beyond empirical comparisons with the corresponding network model, the statistical model can be used to determine the importance of data correlations $\left\{ G_{x}^{(n)} \right\}_n$ to a given task. For example, in classification tasks the network learns to minimise the expectation of a loss function $l(y, t)$ between the network outputs $y$ and class labels $t$. It can be shown that the common choice of
mean-squared error loss, $l_{MSE}(y, t) = \|y-t\|^2$, has an expected value that depends only on the mean $\mu^t_y$ and variance $\Sigma^t_y$ of outputs of each class $t$ \cite{fischer_decomposing_2022} as
\begin{equation}
    L_{\text{MSE}}(\{\mu_y^t, \Sigma_y^t; t\}) = \sum_t p(t) \left( \mathrm{tr}\, \Sigma_y^t + \| \mu_y^t - t \|^2 \right),
\end{equation}
where $p(t)$ is the distribution associated with a particular set of labels $t$. For this loss function, network training and information processing is therefore directly related to understanding how the first- and second-order cumulants of the output arise from the propagation of input data correlation functions. Fischer et al. propose and empirically verify a Gaussian approximation of the statistical model for sufficiently wide networks that use this loss function, in which the hidden layers only pass on the Gaussian part of the statistics $G_{z^l}^{(1)}, G_{z^l}^{(1)}$. While interesting, this is not relevant to the present discussion and we refer readers to \cite{fischer_decomposing_2022} for more information.


\section{CLT: Analytical Solutions for Constrained MLP Weights}
\label{app:analytical solutions}

\subsection{Single Linear Layer}
Networks with a single linear layer and no biases have only two parameters, contained within the weights matrix $W = (w_1, w_2) \in \mathbb{R}^{1 \times 2}$. By comparison with \prettyref{eq:laplace} we can immediately identify $w_0 = w_1 = \frac{1}{\sqrt{2}}$. Besides direct recognition, this result could be obtained by using the cumulant propagation framework to solve for network parameters. However, here we consider an alternative least-squares approach.

We first write the network mapping
\begin{equation}
    y = \begin{pmatrix}
        w_1 & w_2
    \end{pmatrix}
    \begin{pmatrix}
        \xi_1 \\ \xi_2
    \end{pmatrix}
\end{equation}
in keeping with the notation of \prettyref{eq:laplace}, and consider some loss function between network output $y$ and the desired mapping $\frac{1}{\sqrt{2}}(\xi_1 + \xi_2)$
\begin{equation}
    \mathcal{L} = \langle \Big[ w_1 \xi_1 + w_2 \xi_2 - \frac{1}{\sqrt{2}}(\xi_1 + \xi_2) \Big]^2 \rangle.
\end{equation}
Assuming the inputs are both centred i.i.d. variables of variance $\sigma^2$ such that $\langle \xi_1 \xi_2 \rangle = \langle \xi_1 \rangle \langle \xi_2 \rangle$, and defining $\alpha_1 = w_1 -\frac{1}{\sqrt{2}}$, $\alpha_2 = w_2 -\frac{1}{\sqrt{2}}$, we have
\begin{equation}
    \mathcal{L} = (\alpha_1^2 + \alpha_2^2) \sigma^2.
\end{equation}
Minimisation with respect to $\alpha_1, \alpha_2$ directly gives $w_1 = w_2 = \frac{1}{\sqrt{2}}$.

\begin{figure*}[!t]  
    \centering
    \includegraphics[width=\textwidth]{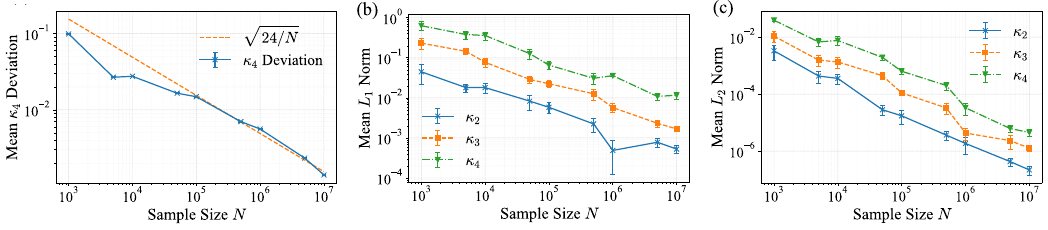}
    \caption{(a) Mean deviation between empirical ground truth and predicted $\kappa_4$ values (using \prettyref{eq:laplace}) as a function of sample size $N$ in the Gaussian task. The theoretical standard deviation $\sqrt{24/N}$ of the sample $\kappa_4$ estimator \cite{joanes_comparing_1998} is overlaid for reference.
    (b), (c) The mean $L_1$ and $L_2$ norms of off-diagonal elements in input tensor cumulants against $N$. $N = 10^6$ was used to balance computational tractability with small $\kappa_4$ error and off-diagonal magnitude.}
    \label{fig:gaussian_dataset_sizes}
\end{figure*}

\begin{figure*}[!b]  
    \centering
    \includegraphics[width=\textwidth]{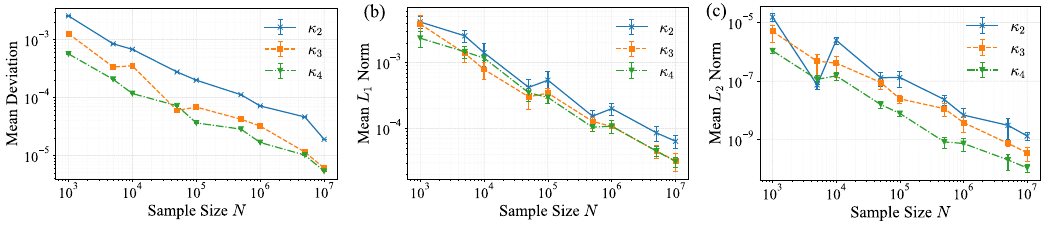}
    \caption{(a) Mean deviation between empirical ground truth and predicted cumulant values (using \prettyref{eq:laplace}) as a function of sample size $N$ in the uniform task, for $\kappa_2, \kappa_3, \kappa_4$. (b), (c) The mean $L_1$ and $L_2$ norms of off-diagonal elements in input tensor cumulants are plotted against $N$. As in the Gaussian task, datasets of size $N = 10^6$ were chosen to balance computational tractability with small $\kappa_4$ error and off-diagonal magnitude.}
    \label{fig:uniform_dataset_sizes}
\end{figure*}

 \begin{figure*}[!t]
  \centering
  \includegraphics[width=\linewidth]{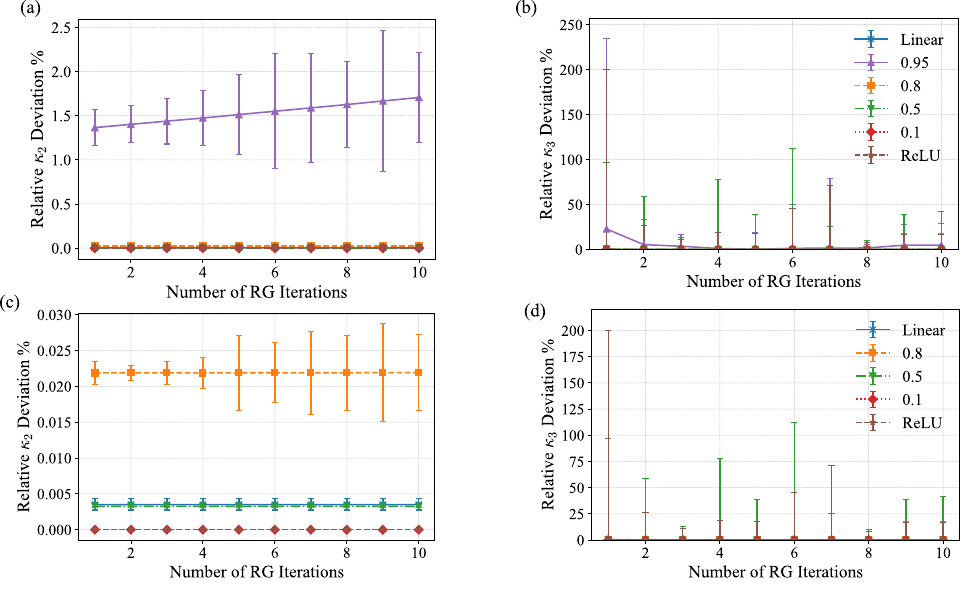}
  \caption{ Relative cumulant deviations as a function of RG steps. (a), (c) are plotted for $\kappa_2$, and (b), (d) for $\kappa_3$. Top panels are plotted with $0.95$ gradient, bottom panels are without. All networks have unconstrained weights, and use the same legend.}
  \label{fig:asymm_leakyrelu_app}
\end{figure*}

 \begin{figure*}[!b]
  \centering
  \includegraphics[width=\linewidth]{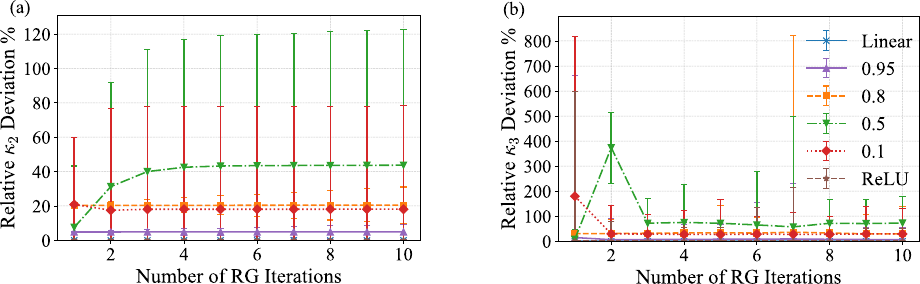}
  \caption{ Relative cumulant deviations as a function of RG steps. (a), (b) are plotted for $\kappa_2$ and $\kappa_3$ respectively. All networks have symmetric weights, and use the same legend.}
  \label{fig:symm_leakyrelu_app}
\end{figure*}

\vspace{1em}
With biases, the network mapping becomes
\begin{equation}
    y = \begin{pmatrix}
        w_1 & w_2
    \end{pmatrix}
    \begin{pmatrix}
        \xi_1 \\ \xi_2
    \end{pmatrix}
    + b
\end{equation}
and we consider the loss function
\begin{equation}
    \mathcal{L} = \langle \Big[w_1\xi_1 + w_2\xi_2 + b - \frac{1}{\sqrt{2}}(\xi_1 + \xi_2)\Big]^2 \rangle.
\end{equation}
Again assuming i.i.d. input variables but now allowing for non-zero mean $\mu = \langle \xi\rangle$, we have $\sigma^2 = \langle \xi^2\rangle - \langle \xi \rangle^2$. Making the same substitutions for $\alpha_1$ and $\alpha_2$, the loss becomes
\begin{align}
    \mathcal{L} &= b^2 + 2b(\alpha_1 + \alpha_2)\mu \nonumber \\
    &\quad + (\alpha_1^2 + \alpha_2^2)\sigma^2 + (\alpha_1 + \alpha_2)^2 \mu^2.
\end{align}
Minimising with respect to $\alpha_1, \alpha_2$ and $b$ yields that $\alpha_1 = \alpha_2 = \alpha$ and $b = -2\alpha \mu$, from which we deduce that
\begin{align}
    w_1 = w_2 &= \frac{1}{\sqrt{2}} \nonumber \\
    b &= 0.
\end{align}

\subsection{Two Linear Layers}

Here we derive \prettyref{eq:linear surface}, using the approach outlined in \prettyref{subsec:analytical weights}. Consider the symmetrically constrained weights
\begin{align}
    W^1 &= \begin{pmatrix}
        w_0 & w_1 \\ w_1 & w_0
    \end{pmatrix} \label{eq:symm W1 app} \\
    W^2 &= \begin{pmatrix}
        w_2 & w_2
    \end{pmatrix} \label{eq:symm W2 app}
\end{align}
and the input cumulants
\begin{align}
    G_{x,\; i}^{(1)} &= \begin{pmatrix}
        \kappa_1 \\ \kappa_1
    \end{pmatrix} \label{eq:G1}\\
    G_{x,\; ij}^{(2)} &= \begin{pmatrix}
        \kappa_2 & 0 \\ 0 & \kappa_2
    \end{pmatrix} \\
    G_{x, \; ijk}^{(3)} &= \begin{cases}
        \kappa_3, & \text{if } i = j = k \\
        0, & \text{otherwise}
\end{cases}, \label{eq:g3 input}
\end{align}
which are diagonal as the inputs are i.i.d. variables. Transforming these according to \prettyref{eq:affine_mean_re} and \prettyref{eq:affine_higher_re}, we find the cumulants of the outputs $y$ to be
\begin{align}
    G_y^{(1)} &= 2\kappa_1w_2(w_0 + w_1) \\
    G_y^{(2)} &= 2 \kappa_2 w_2^2 (w_0 + w_1)^2 \\
    G_y^{(3)} &= 2 \kappa_3 w_2^3 (w_0 + w_1)^3.
\end{align}

From \prettyref{eq:laplace} we can set $G_y^{(1)} = \sqrt{2} G_x^{(1)}$, $G_y^{(2)} = G_x^{(2)}$, and $G_y^{(3)} = \frac{1}{\sqrt{2}} G_x^{(3)}$. Solving this set of simultaneous equations then yields the desired result \prettyref{eq:linear surface}:
\begin{equation}
    w_0 = \frac{1}{w_2 \sqrt{2}} - w_1.
\end{equation}

\vspace{1em}
To introduce biases, define $b = (b_1, b_1)^\top$ for the first layer and $b_2$ for the second. Two additional parameters necessitates transforming cumulants up to fifth order. With $G_x^{(4)}$ and $G_x^{(5)}$ only non-zero when their indices are all the same as in \prettyref{eq:g3 input}, we can perform analogous tensor contractions and use \prettyref{eq:laplace} to find
\begin{align}
    2w_2 \Big[ (w_0 + w_1)\kappa_1 + b_1 \Big] + b_2 &= \kappa_1 \sqrt{2} \\
    2w_2^2 (w_0 + w_1)^2 &= 1 \\
    2w_2^3 (w_0 + w_1)^3 &= \frac{1}{\sqrt{2}} \\
    2w_2^4 (w_0 + w_1)^4 &= \frac{1}{2} \\
    2w_2^5 (w_0 + w_1)^5 &= \frac{1}{2\sqrt{2}}.
\end{align}
Solving this set of equations yields \prettyref{eq:subspace with biases}
\begin{align}
    w_0 &= \frac{1}{w_2 \sqrt{2}} - w_1 \\
    b_2 &= -2w_2 b_1.
\end{align}
All tensor contractions were checked using the SymPy library on Python (see \href{https://docs.sympy.org/latest/index.html}{here} for more information).

\subsection{Quadratic Nonlinearity}

Here we show how the inconsistency in \prettyref{eq:inconsistency} arises from cumulant propagation in networks with two linear layers as above, with the same weights, no biases, and an additional quadratic nonlinearity $\phi(z) = z + \alpha z^2$.

Proceeding layerwise, after the first affine transformation the first two cumulants are
\begin{align}
    G_{z^1}^{(1)} &= (w_0 + w_1) \kappa_1 \begin{pmatrix}
        1 \\ 1
    \end{pmatrix} \\
    G_{z^1}^{(2)} &= \kappa_2 \begin{pmatrix}
        w_0^2 + w_1^2 & 2w_0 w_1 \\ 2w_0 w_1 & w_0^2 + w_1^2
    \end{pmatrix}.
\end{align}
Applying the interaction function in \prettyref{eq:mean interaction function}, after the nonlinear activation the first cumulant becomes
\begin{equation}
    G_{y^1, \;i}^{(1)} = (w_0 + w_1)\kappa_1 + \alpha \kappa_1^2 (w_0 + w_1)^2 + \alpha \kappa_2(w_0^2 + w_1^2),
\end{equation}
which is the same for both elements $G_{y^1, \;1}^{(1)}, G_{y^1, \;2}^{(1)}$. After the second affine transformation, comparison with the expected cumulant evolution in \prettyref{eq:laplace} yields
\begin{align}
    \kappa_1 \sqrt{2} &= 2 w_2 \Big[ (w_0 + w_1)\kappa_1 + \alpha \kappa_1^2 (w_0 + w_1)^2 \nonumber \\
    &\quad + \alpha \kappa_2 (w_0^2 + w_1^2) \Big].
\end{align}
Successful learning of the transformation means that this relationship should hold for any values of $\kappa_1$ and $\kappa_2$. Thus, comparing coefficients leads to the equations
\begin{align}
    2 w_2 (w_0 + w_1) &= \sqrt{2} \nonumber \\
    2w_2 \alpha (w_0 + w_1)^2 &= 0 \nonumber \\
    2w_2 \alpha (w_0^2 + w_1^2) &= 0,
\end{align}
and it is clear from the latter two that this set has a contradiction.

\vspace{1em}
Weights could alternately be constrained as
\begin{align}
    W^1 &= \begin{pmatrix}
        w_0 & w_0 \\ w_1 & w_1
    \end{pmatrix} \\
    W^2 &= \begin{pmatrix}
        w_2 & w_2
    \end{pmatrix},
\end{align}
as these satisfy the constraint $W^2 W^1 P = W^2 W^1$ which preserves the required $S_2$ symmetry. This 'columns-same' approach is less expressive as it collapses the representation to a one-dimensional subspace of parameters, whereas weights that are symmetric across the diagonal preserve the two-dimensionality of the hidden representation space. Nonetheless, in this case we draw the same conclusions. Following the exact same steps leads to the equations
\begin{align}
    2w_2 (w_0 + w_1) &= \sqrt{2} \nonumber\\
    4 w_2 \alpha (w_0^2 + w_1^2) &= 0 \nonumber\\
    2 w_2 \alpha (w_0^2 + w_1^2) &= 0,
\end{align}
which have no real solution.

\section{Experimental Setup}
\label{app:exp setup}

This appendix contains key hyperparameters and training details used to implement network models as in \prettyref{subsec:MLP training setup} and \prettyref{subsec:GNN architecture}. We also include empirical results justifying our choice of data set size $n = 10^6$ for both the Gaussian and uniform task in \prettyref{fig:gaussian_dataset_sizes} and \prettyref{fig:uniform_dataset_sizes}.

\subsection{MLPs}

Before training, network parameters were all randomly initialised from i.i.d. centred Gaussians $W_{ij}^{\text{i.i.d.}} \sim \mathcal{N}(0, \sigma_w^2 / N_{l-1})$ and $b_i^{\text{i.i.d.}} \sim \mathcal{N}(0, \sigma_b^2)$. The variance scaling $\sigma_w^2 / N_{l-1}$ is such that the covariance of $z^l$ is independent of the layer width, as inspired by both Glorot \& Bengio and He et al \cite{glorot_understanding_2010, he_delving_2015}.

Variances $\sigma_w^2 = \sigma_b^2 = 0.75$ were used to initialise network parameters $\theta$. Networks were all trained using the empirical MSE loss per data batch $\{(x^{(b)}, t^{(b)})\}_b$:
\begin{equation}
    R_{\text{emp}, \text{MSE}}(\theta) = \frac{1}{B} \sum_{b=1}^{B}
    \|g(x^{(b)};\theta) - t^{(b)}\|^2.
\end{equation}
Training was limited to two epochs to investigate generalisation before models overfitted to specific tasks. Batch size $B$ was set to 64. \textsc{ADAM} \cite{kingma_adam_2017, loshchilov_decoupled_2019} was used for optimisation with learning rate $10^{-3}$, momenta $\beta_1 = 0.9$ and $\beta_2 = 0.999$, $\epsilon = 10^{-8}$, and weight decay $\lambda=0$. The choice of optimiser doesn't affect any previous theoretical results \cite{fischer_decomposing_2022}.

\subsection{GNNs}


Parameters $\theta$ were initialised from i.i.d. Gaussians with $\sigma_w^2 = \sigma_b^2 = 0.75$ as for MLPs. Networks were trained using the empirical MSE loss per data batch, with batch size $64$, and \textsc{Adam} was used for optimisation with the same hyperparameters as above. While weights were not actively constrained to be symmetric, we added a soft symmetry regularisation term to the loss to improve training stability. A weight norm penalty was also used to prevent weights from collapsing to vanishingly small values. Due to the directed nature of input graphs, networks were trained solely on the embeddings of target nodes. As for MLPs, all data sets were of size $n=10^6$ and shuffled duplicates of network outputs were used as inputs for successive RG steps to ensure i.i.d. conditions. 

\section{MLP Uniform Task: Leaky ReLU Results}
\label{app:extra MLP results}

\prettyref{fig:asymm_leakyrelu_main} and \prettyref{fig:symm_leakyrelu_main} in the main text illustrate the key findings of our experiments using the leaky ReLU activation. For completeness, we include here the relevant cumulant deviation graphs for $\kappa_2$ and $\kappa_3$ in both cases. \prettyref{fig:asymm_leakyrelu_app} contains results for asymmetric weights, while \prettyref{fig:symm_leakyrelu_app} contains those for symmetrically-constrained weights.

\end{appendices}

\bibliographystyle{unsrt}
\bibliography{refs, Part_III_Project}

\end{document}